\documentclass[conference]{IEEEtran}
\IEEEoverridecommandlockouts
\usepackage{cite}
\usepackage{amsmath,amssymb,amsfonts}
\usepackage{algorithmic}
\usepackage{graphicx}
\usepackage{textcomp}
\usepackage{lipsum}
\usepackage{mathrsfs}
\usepackage{color}
\usepackage{hyperref}
\usepackage{tabularx}
\usepackage{multirow}
\usepackage{booktabs}
\usepackage{cite}
\usepackage{float}
\usepackage{subfig}
\usepackage{caption}
\usepackage{tikz}
\graphicspath{ {./figs/} }
\DeclareGraphicsExtensions{.pdf,.jpeg,.png}

\def\BibTeX{{\rm B\kern-.05em{\sc i\kern-.025em b}\kern-.08em
    T\kern-.1667em\lower.7ex\hbox{E}\kern-.125emX}}

\makeatletter
\newcommand{\linebreakand}{%
  \end{@IEEEauthorhalign}
  \hfill\mbox{}\par
  \mbox{}\hfill\begin{@IEEEauthorhalign}
}
\makeatother

\begin{document}

\title{
    GET-DIPP: Graph-Embedded Transformer for Differentiable Integrated Prediction and Planning
    
}

\author{
    \IEEEauthorblockN{1\textsuperscript{st} Jiawei Sun}
    \IEEEauthorblockA{
        \textit{National University of Singapore} \\
        Singapore \\
      sunjiawei@u.nus.edu}
    \and
    \IEEEauthorblockN{2\textsuperscript{nd} Chengran Yuan}
    \IEEEauthorblockA{
        \textit{National University of Singapore} \\
        Singapore \\
        chengran.yuan@u.nus.edu}
    \and
    \IEEEauthorblockN{3\textsuperscript{rd} Shuo Sun}
    \IEEEauthorblockA{
        \textit{National University of Singapore} \\
        Singapore \\
        shuo.sun@u.nus.edu}
    \linebreakand 
    \IEEEauthorblockN{4\textsuperscript{th} Zhiyang Liu}
    \IEEEauthorblockA{
        \textit{National University of Singapore} \\
        Singapore \\
        zhiyang@u.nus.edu}
    \and
    \IEEEauthorblockN{5\textsuperscript{th} Terence Goh}
    \IEEEauthorblockA{
        \textit{Yinson} \\
        Singapore \\
        terence.goh@yinson.com}
    \and
    \IEEEauthorblockN{6\textsuperscript{th} Anthony Wong}
    \IEEEauthorblockA{
        \textit{MooVita} \\
        Singapore  \\
        anthonywong@moovita.com}
    \linebreakand 
    \IEEEauthorblockN{7\textsuperscript{th} Keng Peng Tee}
    \IEEEauthorblockA{
        \textit{MooVita} \\
        Singapore  \\
        kptee@moovita.com}
    \and
    \IEEEauthorblockN{8\textsuperscript{th} Marcelo H. Ang Jr.}
    \IEEEauthorblockA{
        \textit{National University of Singapore}\\
        Singapore \\
        mpeangh@nus.edu.sg}
}

\maketitle

\begin{abstract}
Accurately predicting interactive road agents' future trajectories and planning a socially compliant and human-like trajectory accordingly are important for autonomous vehicles. In this paper, we propose a planning-centric prediction neural network, which takes surrounding agents' historical states and map context information as input, and outputs the joint multi-modal prediction trajectories for surrounding agents, as well as a sequence of control commands for the ego vehicle by imitation learning. An agent-agent interaction module along the time axis is proposed in our network architecture to better comprehend the relationship among all the other intelligent agents on the road. To incorporate the map's topological information, a Dynamic Graph Convolutional Neural Network (DGCNN) is employed to process the road network topology. Besides, the whole architecture can serve as a backbone for the Differentiable Integrated motion Prediction with Planning (DIPP) method by providing accurate prediction results and initial planning commands. Experiments are conducted on real-world datasets to demonstrate the improvements made by our proposed method in both planning and prediction accuracy compared to the previous state-of-the-art methods.%
\end{abstract} 


\section{Introduction}

In the past few years, research on Autonomous Vehicles (AVs) has been gaining increasing popularity in both academia and industry. Among these works, studies on prediction and planning have risen steadily due to the growing industry demand and ample mature solutions for upstream tasks in AV systems. Benefiting from the developed software and hardware of perception, multi-modality scenario information can be effortlessly acquired and then fed into the downstream prediction networks. Recent works \cite{gaoVectorNetEncodingHD2020}, \cite{ balakrishnanvaradarajanMultiPathEfficientInformation2021}, \cite{junruguDenseTNTEndtoendTrajectory2021} based on transformers exploited the scene context via vector representation. However, most of these methods only forecast the agents' future trajectories without considering the underlying interactions among heterogeneous road users, which may result in non-convergent trajectories. To tackle these issues and achieve accurate multi-agent prediction, some recent works \cite{xiaoyumoMultiAgentTrajectoryPrediction2022}, \cite{jiaMultiAgentTrajectoryPrediction2022} were proposed to model the interactions among agents. 

Progress in motion prediction exerts a positive effect on trajectory planning, which is a long-developed and vibrant research area. Conventional approaches in trajectory planning employ a wide variety of techniques such as sampling \cite{aoudeThreatawarePathPlanning2010}, \cite{karamanAnytimeMotionPlanning2011}, \cite{Sun_FISS_A_Trajectory_2022}, graph search  \cite{bohrenLittleBenBen2008}, \cite{kushleyevTimeboundedLatticeEfficient2009} optimization \cite{guOnRoadMotionPlanning2012}, etc. Due to the limited generalization ability and non-human-like driving behaviors inherited in conventional methods, imitation learning \cite{bansal2018chauffeurnet},\cite{qinDeepImitationLearning2021a},\cite{weiss2022way} and reinforcement learning \cite{li2021hierarchical},\cite{huang2022efficient} methods are expected to enable vehicles to drive naturally in complex traffic scenarios. Motion prediction and planning are two tightly-coupled components within the AV stack, and some recent attempts were made to perform two tasks jointly. There are two main categories of these exploratory works, selecting the optimal trajectory after evaluation of the set of candidates \cite{liu2021deep},\cite{cui2021lookout} and predicting the cost map within a short horizon before planning \cite{zeng2019end},\cite{hu2021safe}, both of which suffer from expensive computation and low efficiency. Combining prediction and planning in an end-to-end model \cite{ngiam2021scene} can contribute to increased accuracy but not guaranteed safety. The recent work DIPP \cite{huang2022differentiable} proposed a differentiable non-linear optimizer after prediction and imitation planning module which explicitly considers dynamic constraints and traffic rules, enhancing performance and safety. However, DIPP's backbone only consists of the agent-agent and agent-map interaction relationship, neglecting the topological relationship between lanes and crosswalks. Considering this drawback, we propose our DGCNN and transformer-based network structure.

The main contributions of this paper can be summarized as follows:
\begin{enumerate}
    \item We propose an end-to-end planning-centric prediction neural network that can output both neighbor agents' prediction trajectories and ego vehicle's future control variables.
    \item We introduce a transformer-based agent-agent interaction module along the time axis to better model the relationship between agents and employ the DGCNN\cite{wang2019dynamic} module to extract the topological information from the map.  
    \item A novel proxy way-points strategy is proposed to reduce the memory consumption of the DGCNN network.
\end{enumerate}

\begin{figure*}[ht]
    \centering
    \includegraphics[width=\linewidth]{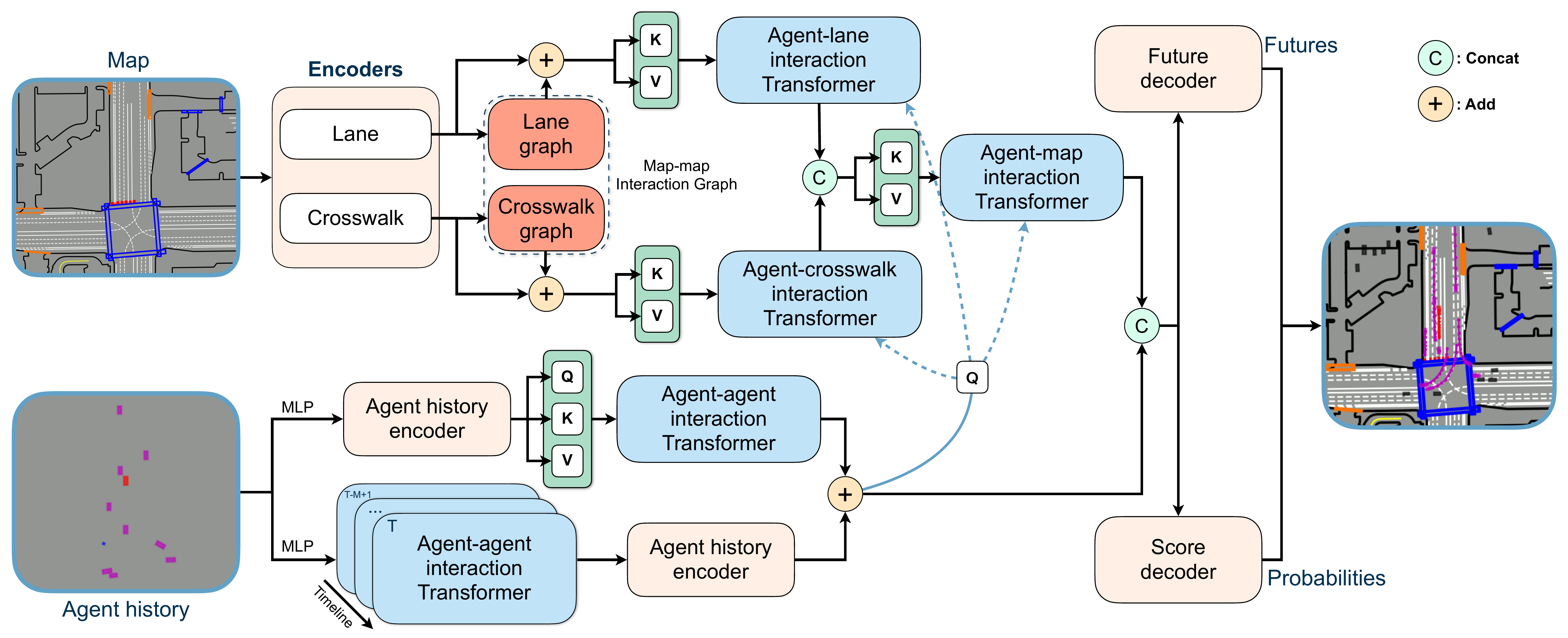}
    \caption{Overview of the proposed GET-DIPP network architecture.}
    \label{figure-ours}
\end{figure*}

\section{Methodology}

\subsection{Problem Definition}
A given driving scenario consists of two main parts: interactive agents and map contexts. All the interactive agents are denoted as $\{{D}_{i}|i=0, \dots, K\}$, where $D_{0}$ denotes ego vehicles and $D_{i}$ denotes surrounding road participants such as pedestrians, cyclists, and other vehicles. The map context information at each time-step $t$ is represented as $C_{t}$. The motion prediction task aims to predict the future states of the surrounding agents based on their historical states and corresponding local map context information. All agents' historical states are discrete in time, and for one agent $D_{i}$, its past $M$ historical states can be denoted as $D_{i}^{t-M+1:t}$ and its forecast ${N}$ future trajectories can be denoted as $\hat{D}_{i}^{t+1:t+N}$. The predictor model ${P}$ takes all agents' history states and current time step map context information as input and outputs $X$ different future trajectories for each surrounding agent and the possibility $p$ for each future trajectory. Besides, the model will also serve as an imitation learning planner to generate $X$ different control variables ${u^{t+1:t+N}}$for the ego vehicle. Then, the problem is mathematically abstracted as:
\begin{equation}
\label{eq1}
    \begin{aligned}
    P(D_{0}^{t-M+1:t}, \dots , &D_{K}^{t-M+1:t}|C_{t}) \\[8pt]
    &= \left\{ 
    \begin{array}{ll}
    \{u_{0,j}^{t+1:t+N},{p}_{j}\}_{j=1}^{X}, {i=0}\\[8pt]
    \{\hat{D}_{i,j}^{t+1:t+N},{p}_{j}\}_{j=1}^{X}, {i\neq0}.
    \end{array}\right.
    \end{aligned}
\end{equation}

\subsection{Network Framework}
\textbf{Input representation}. We follow the input strategy described in \cite{huang2022differentiable}, feeding all agents' historical states and the current map context information into the network. Since the historical states are discrete in the time domain, a two-dimensional tensor $[M,F]$ can be used, where $M$ denotes the historical state's length and $F$ denotes the feature dimension. The feature dimension consists of 2D coordinates, velocity, heading, and the size of each bounding box. The ego vehicle's $K$ nearest (in Euclidean distance) agents are taken into consideration in the prediction task. As another input to the network, the map context information consists of vectorized lanes' waypoints and crosswalks' waypoints. Each lane waypoint feature encompasses lane boundaries, lane center line, speed limit, traffic light, stop point, etc. Each crosswalk waypoint feature encompasses the 2D position and heading angle of a crosswalk waypoint. For each agent, 6 nearest lanes and 4 nearest crosswalks are combined together as its local map context information fed into the network. 

\indent\textbf{Lane Encoder} The input of this block is lane context information, which is essentially a series of waypoint features ${[L, F_{1}]}$, where $L$ denotes the waypoint number and $F_{1}$ denotes the original feature dimension. The $F_{1}$-dimensional feature consists of the left and right boundaries (2D coordinates and heading angles), the centerline (2D coordinates and heading angles), speed limit, left boundary type, right boundary type, centerline type, traffic light states, stop signs, and interpolating signs. For numerical features like the boundaries' 2D coordinates, a Fully Connected($FC$) layer is utilized for encoding it into high-dimensional feature space. For integer-type features like traffic light states, an embedding layer is used to extract discrete label features. Subsequently, all the encoded integer-type features are directly summed up and then concatenated with all the encoded numeric features to form a $F_{2}$-dimensional encoded waypoint feature. Finally, a multi-layer perceptron$(MLP)$ with $ReLU$ activation function is utilized to map the waypoint feature to an appropriate dimension. The output of this lane encoder block is an encoded waypoint feature of the shape ${[L,F]}$.

\indent\textbf{Crosswalk Encoder}. The input of this block is all the waypoints$[L,3]$ that encompass a pedestrian walk, where $L$ denotes the number of points, and $3$ denotes the 2D positions and heading angle. A $(MLP)$ with $ReLU$ function block is utilized for encoding the point feature into high dimensional space. The output of this block is the encoded crosswalk points feature of the shape ${[L,F]}$.

\indent\textbf{Agent History Encoder}. All agents' historical states are encoded by a shared two-layer Long Short Term Memory($LSTM$) network. For each agent, only the last(current) time step's output feature vector of the ($LSTM$) network is utilized as the encoded output. All agents' encoded outputs are stacked together as the final output.

\indent\textbf{Agent-Agent Interaction}. A shared two-layer multi-head self-attention encoder is used to extract the relationships among interactive agents. Two encoding strategies are adopted in parallel to model the agent-agent interaction at different levels. The first strategy is to encode the agent-agent interaction at every historical time step in which the input queries, keys, and values are the corresponding agents' features gone through a ${FC}$ layer to align with the input dimension. Subsequently, the output is fed into the agent history encoder. The second strategy is to do an agent history encoder on the time axis and then feed the output into the agent-agent interaction block. In this case, the queries, keys, and values are encoded agents' historical state features. The self-attention mechanism for each head can be written as:
\begin{equation}
Attention(Q,K,V)=softmax(\frac{QK^{T}}{\sqrt{d_{k}}}V)
\end{equation}

\indent\textbf{Map-Map Interaction}. This block applies a Dynamic Graph Convolutional Neural Network (DGCNN) structure \cite{wang2019dynamic} to extract maps' topological information. Consider a $F$-dimensional waypoint feature set, denoted by $\Omega=\{P_{1},P_{2},\dots,P_{L}\} \subseteq \mathbb{R}^{F}$ and in our case $F=256$. A directed graph $G=\{V, E\}$ is utilized to represent local waypoint relationships where waypoints as vertices $V=\{1,\dots, L\}$ and their interactions as edges $E \subseteq \{V\times V\}$. The graph, including self-loops, is established by finding the k-nearest neighbor of each vertex in feature space $\mathbb{R}^{F}$. The edge features are derived by $e_{ij}=f_{\Theta}(P_{i},P_{i}-P_j{})$, where $f_{\Theta}:\mathbb{R}^{F}\times\mathbb{R}^{F}\to \mathbb{R}^{F^{'}} $ is a Multi-Layer Perceptron(MLP) with $ReLU$ activation function. Finally, a channel-wise max-pooling operation is applied on the edge features emitting from each vertex to gain new vertex features. This process can be described as:
\begin{equation}
P_{i^{'}}=maxpooling(f_{\Theta}(P_{i},P_{i}-P_j{}))\subseteq\mathbb{R}^{F^{'}}
\end{equation}

\begin{figure*}[ht]
    \centering
    \subfloat[Lane Graph]{\includegraphics[height=0.4\linewidth]{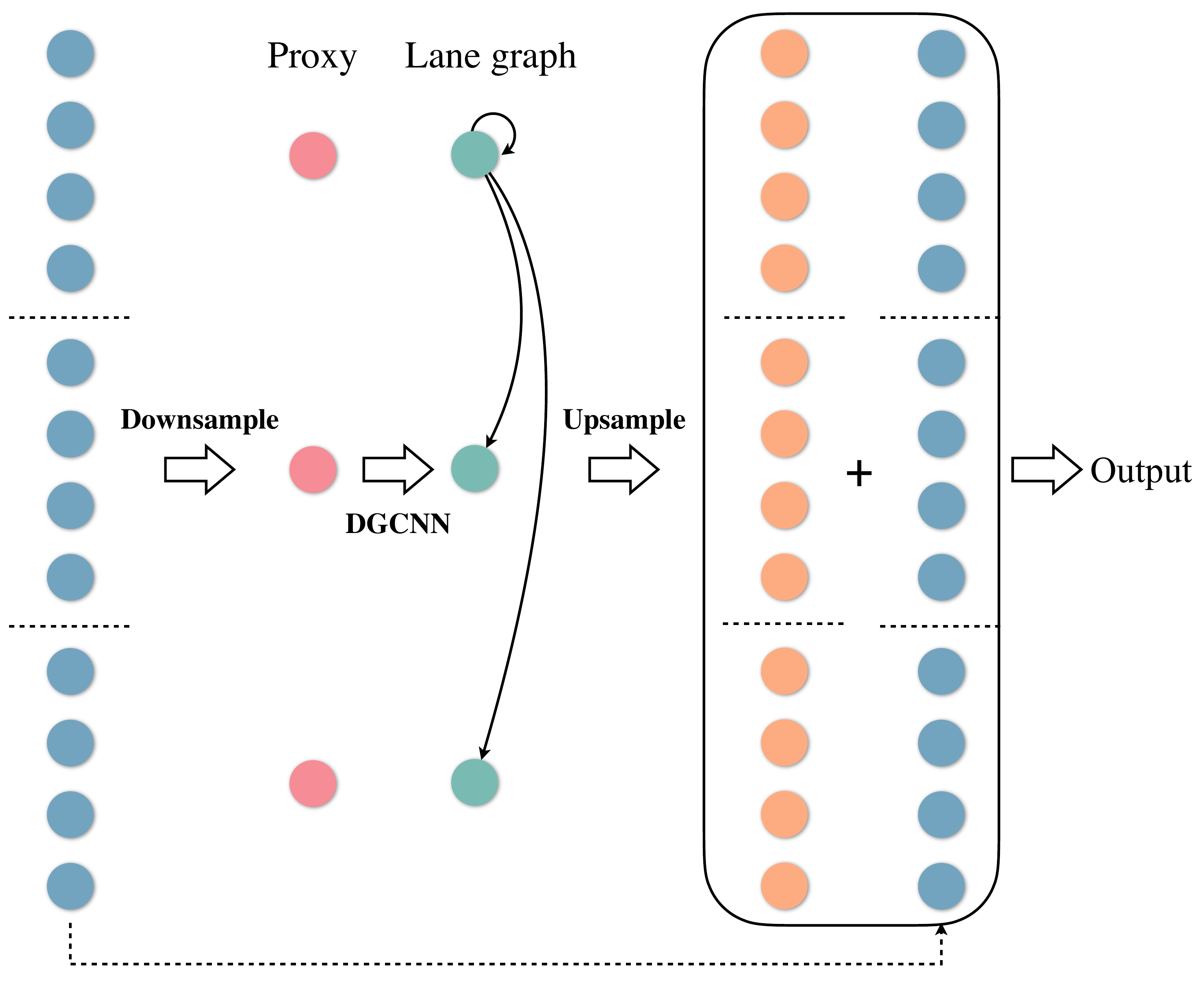}}
    \label{subfig-lane}
    \hfill
    \subfloat[Crosswalk Graph]{\includegraphics[height=0.4\linewidth]{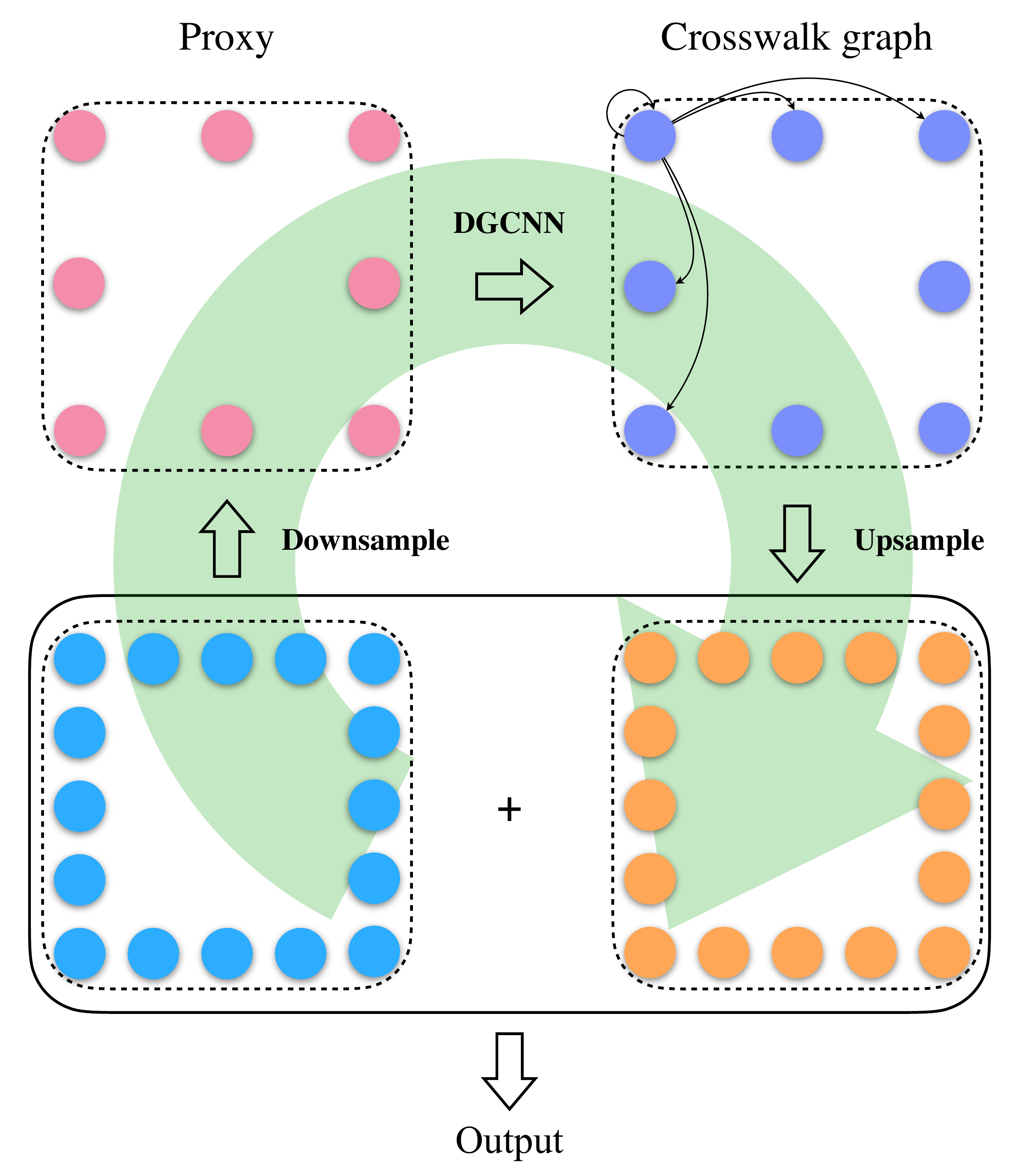}}
    \label{subfig-crosswalk}
    \caption{Illustration of the lane graph (a) and crosswalk graph (b). The encoded waypoints (in blue) first go through a down-sampling process to assemble the proxy waypoints, followed by a DGCNN network to incorporate neighborhood information. Finally, the output proxy waypoints feature will be duplicated and added with a shortcut from the original waypoint features.}
    \label{figure-encoders}
\end{figure*}

In the lane graph Fig.\ref{figure-encoders}(a), considering the huge number of waypoints in the scene, a proxy waypoint strategy is applied to reduce memory consumption. The waypoints set is equally split into several subsets $\{\{P_{1},\dots,P_{l},\},\dots,\{P_{L-l+1},\dots,P_{L},\}\}$. The average of each subset is computed to assemble a proxy waypoint, which can be viewed as a type of down-sampling operation. Subsequently, all the proxy waypoints are fed into a $DGCNN$ network. Finally, the output proxy waypoints feature will be duplicated $l$ times and added with a shortcut from the original waypoint features.
Crosswalk graph Fig.\ref{figure-encoders}(b) block adopts the same strategy illustrated above. The memory consumption in this part will decrease from $O(L\times L)$ to $O(\frac{L}{l}\times \frac{L}{l})$. Map-map interaction incorporates each waypoint's local neighborhood information, thus giving the network topological information to achieve better performance.

\indent\textbf{Agent-Map Interaction}. A hierarchical cross-attention transformer block is utilized to model the interaction between agents to map.
Firstly, each agent's encoded interaction feature will perform cross-attention with one of its surrounding encoded map vectors (encoded lanes or crosswalks) iteratively. The query is the agent's interaction feature, while the key and value are one map vector. Since one map vector is a sequence of encoded waypoints with topological information, the output can be viewed as the agent's attention on the map waypoints. Stacking all the outputs together gives a sequence of map attention vectors (agent-lane and agent-crosswalk). Then, three cross-attention encoder modules are arranged in parallel to model the multi-modal relationships between agent and map attention vectors. The query used in this step is still the agent's interaction feature whereas the key and value are map attention vectors. The final output is the multi-modal relationships between one agent and its surrounding map context. Then, the above process is iteratively repeated for all the agents and stacking the agent-map outputs together to give the multi-modal agent-map interaction results.

\indent\textbf{Decoder}. For each agent, the agent-agent interaction vector is duplicated and concatenated with the multi-modal agent-map interaction vectors. Then, for the ego vehicle, a $MLP$ block is used to decode a series of future acceleration and steering angle values. Given the discrete time series of control variables, a kinematic bicycle model is used to project the planned trajectories forward. For the surrounding agents, a shared $MLP$ block is used to decode their possible future trajectories. To evaluate the possible predicted trajectories, the agent-agent and agent-map interactions are first aggregated by a max-pooling operation, followed by another $MLP$ block to give out the probabilities for the joint future trajectories.

\subsection{Loss Function}. The entire neural network is trained in an end-to-end manner. The total loss encompasses four parts: trajectory regression loss, score loss and imitation loss, where the imitation loss can be split into planning Average Displacement Error (ADE) loss and planning Final Displacement Error (FDE) loss.
\begin{equation}
    \mathcal{L}_{total} = \lambda _{1}\mathcal{L}_{predict}+\lambda _{2}\mathcal{L}_{score}+\lambda _{3}\mathcal{L}_{ADE}+\lambda _{4}\mathcal{L}_{FDE}
\end{equation}
where $\lambda_{1,2,3,4}$ are hyper-parameters to balance the different loss terms.

The prediction regression loss is defined as the smooth $L1$ loss between the ground truth trajectory($\pi^{*}$) and the predicted trajectory($\pi$) that is closest to ground truth:
\begin{equation}
    \mathcal{L}_{prediction} = smooth\mathcal{L}_{1}(\pi - \pi^{*})
\end{equation}

The score loss is defined as the cross entropy loss between the predicted different futures' possibilities and ground truth:
\begin{equation}
    \mathcal{L}_{score} = \mathcal{L}_{cross\_entropy}(p, p^{*})
\end{equation}

The planning FDE loss is defined as the Euclidean distance between the ego's final coordinate and the ground truth:
\begin{equation}
    \mathcal{L}_{FDE} = \mathcal{L}_{2}(\pi_{ego, t=T} - \pi^{*}_{ego, t=T})
\end{equation}

The planning ADE loss is defined as the average Euclidean displacement between the planned trajectory and ground truth:
\begin{equation}
    \mathcal{L}_{ADE} = 
        \frac{ 
            \sum_{t=0}^{t=T}(\mathcal{L}_{2}(\pi_{ego, t} - \pi^{*}_{ego, t})) \times \Delta{t}
        }{T}
\end{equation}

\begin{figure*}[ht]
    \centering
    \includegraphics[width=0.99\linewidth]{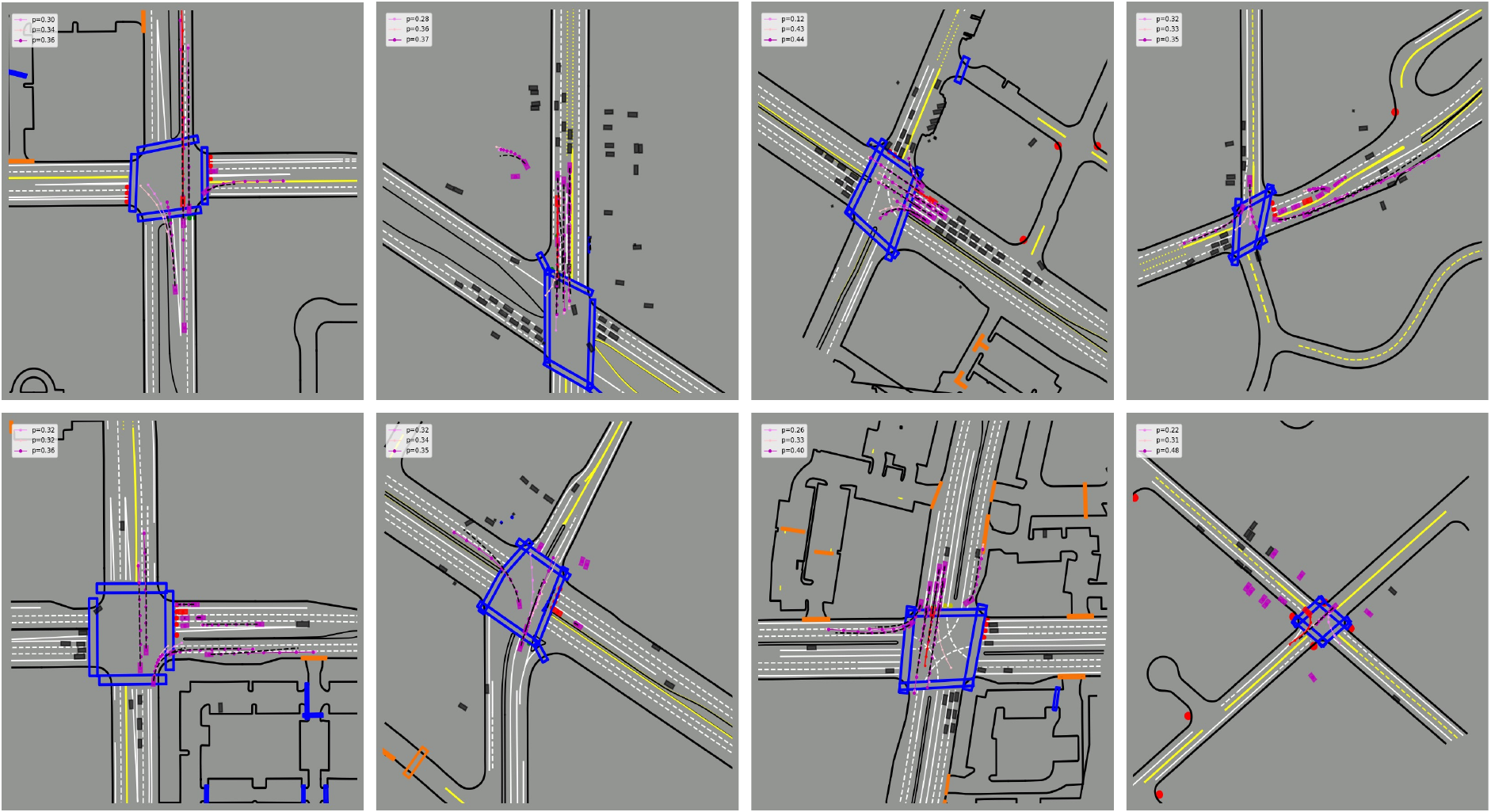}
    \caption{Qualitative results for multi-modal motion predictions. The red, magenta and black boxes represent the ego vehicle, interactive agents, and other vehicles respectively. The magenta, pink, and violet lines are predicted trajectories. The red line stands for the planned trajectory for the ego vehicle. The black lines are the ground truth for all road participants.}
    \label{figure-ours-pred}
\end{figure*}

\section{Experiments}
\subsection{Dataset}
The proposed network structure is trained and validated on Waymo open motion dataset \cite{ettinger2021large}, which contains 104,000 diverse urban driving scenarios(each 20 seconds duration at 10$Hz$) and comprehensive map and agents information. 14696 scenarios are chosen from the dataset, where 84.4\% are used for training, 12.1\% are used for testing, and 3.5\% for testing. Each scenario is split into several 7-second frames, including 2-second history, 5-second future, and corresponding map information. The total number of frames for training is 173438, 24935 for validation, and 7137 for testing.

\subsection{Evaluation}
The prediction performance of the proposed network is evaluated in the test dataset. For planning performance, an extra log-replay simulator is constructed, where at each time step, only the first control variable ($a_0$,$\delta_0$) is taken by the ego vehicle and its state will be updated based on the bicycle model, while other road participants will move as their ground truth trajectories in the dataset. Collision rate, off-route rate, progress, average longitudinal acceleration, average jerk, average lateral acceleration, prediction ADE, prediction FDE, and planning error are the main metrics used to judge the model.

\subsection{Comparison baselines}
The proposed network structure is compared with the DIPP's backbone network in the following three aspects:
\begin{itemize}
    \item \textbf{Imitation learning with prediction sub-task:} the network is trained in an end-to-end fashion to infer both imitated motion planning commands and the predicted trajectories of the surrounding agents. The overall architecture of the method is shown in Fig. \ref{figure-ours}.
    \item \textbf{Imitation learning with prediction sub-task followed by another motion planner:} We take the network's output planning commands as initial planning and feed them into the same differentiable non-linear optimization planning module as DIPP together with the best prediction result. This one is like a two-stage planning method. 
    \item \textbf{Integrated prediction with motion planner:} we train our proposed network structure with the same differentiable non-linear optimization planning module as DIPP in an end-to-end fashion.
\end{itemize}

\subsection{Training} 
All of the experiments are conducted on a workstation equipped with one NVIDIA TITAN RTX graphic card. The batch size used in training is $32$, together with the Adam optimizer with an initial learning rate of $2e-4$, reduced by half after every $4$ epochs.
The dimension of encoded waypoints, crosswalk points, and encoded agents' historical state is set to $256$. The number of heads for the agent-agent, agent-lane, and agent-crosswalk interaction transformer is $8$, and for the agent-map interaction, the transformer is $4$. For the lane graph and crosswalk graph, a two-layer DGCNN block is used, and the number of $K$-nearest neighbor is set to $20$. The loss weights are set as $\lambda_{1} = 0.5, \lambda_{2} = \lambda_{3} = \lambda_{4} =1$.

\begin{figure*}[ht]
    \centering
    \includegraphics[width=0.99\linewidth]{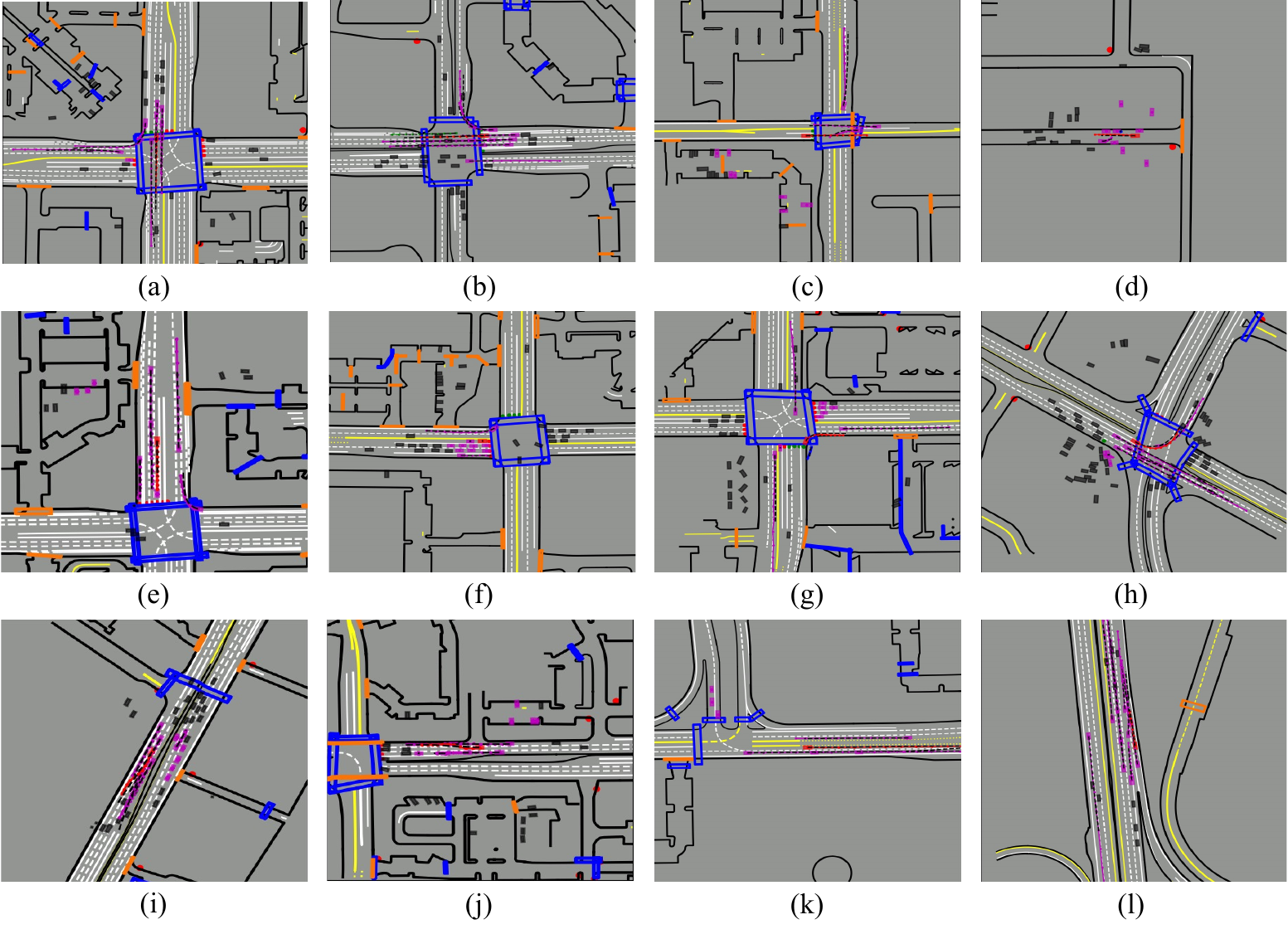}
    \caption{Qualitative results of the planned ego trajectories by our proposed method. The red, magenta and black boxes represent the ego vehicle, interactive agents, and the rest of the vehicles on the road respectively. The magenta line is the top-1 predicted trajectory. The red line stands for the planned trajectory for the ego vehicle. The black lines are the ground truth trajectories for all road participants. }
    \label{figure-ours-plan}
\end{figure*}

\section{Results and Discussions}

\subsection{Multi-modal Predictions Results}
To demonstrate the effectiveness of our planning-centric prediction model, 8 representative scenarios, as well as the different possible prediction results, are visualized in Fig. \ref{figure-ours-pred}. For the straight cruising vehicles, the 3 predicted trajectories are nearly converged to one trajectory, with slight differences in length. The predicted possible trajectories will diverge in different directions when the vehicle confronts an intersection. In the case of waiting for the traffic light, all the predictions stay still. And the trajectory closest to the ground truth is assigned with the highest probability by our network. 

\begin{table}[h]
\caption{Quantitative results for multi-modal motion predictions.}
    \begin{tabular}{c|c|ccc|ccc|ccccc}
    \toprule
          & \multirow{2}[2]{*}{Method} & \multicolumn{2}{c}{Prediction error(m)} \\
          &         & ADE   & FDE \\
    \midrule
    \multirow{3}[2]{*}{DIPP\cite{huang2022differentiable}} & IL+Prediction  & 0.773 & 1.916 \\
          & IL+Prediction+Planner       & 0.766 & 1.908 \\
          & Integrated Prediction with Planning       & 0.740 & 1.814 \\
    \midrule
    \multirow{3}[2]{*}{Ours} & IL+Prediction & 0.733 & 1.850 \\
          & IL+Prediction+Planner  & 0.730 & 1.845 \\
          & Integrated Prediction with Planning & \textbf{0.712} & \textbf{1.808} \\
    \bottomrule
    \end{tabular}%
  \label{table-1}%
\end{table}%

 Table \ref{table-1} shows the quantitative results for our prediction results compared to previous works. The prediction ADE and FDE are improved by $5.17$\% and $3.44$\% respectively compared against the first baseline, suggesting that our network is able to better contemplate the relationship between interactive agents and incorporate rich map's topological information. It is found that training the backbone network together with the differential motion planner will help to achieve better prediction accuracy. In the third benchmark experiment, where our backbone is trained together with the differentiable motion planner in an end-to-end fashion, our prediction ADE and FDE are improved by $3.7\%$ and $0.33\%$. Generally, the results show that our model can provide more accurate multi-modal joint prediction results for the ego vehicle's surrounding agents compared to the original backbone for DIPP.

\begin{table*}[hbtp]
  \centering
  \caption{Quantitative results for planning.}
    \begin{tabular}{c|c|ccc|ccc|ccccc}
    \toprule
          & \multirow{2}[2]{*}{Method} & Collision & Off route & Progress  & Acc.  & Jerk  & Lat. Acc.  & \multicolumn{3}{c}{Position error(m)}  \\
          &       &  (\%) &  (\%) & (m)   & (m/s2) & (m/s2) & (m/s2) & @3s   & @5s   & @10s   \\
    \midrule
    \multirow{3}[2]{*}{DIPP\cite{huang2022differentiable}} & IL+Prediction & 42    & 58    & 11.05 & 1.345 & 1.977 & 0.813 & 11.03 & 19.13 & 42.25 & \\
          & IL+Prediction+Planner & 7     & \textbf{0} & 76.28 & 0.638 & 1.313 & 0.058 & 1.869 & 4.182 & 10.17  \\
          & Integrated Prediction with Planning & 5     & \textbf{0} & \textbf{77.57} & 0.624 & 1.209 & 0.069 & 1.726 & 3.913 & 9.365  \\
    \midrule
    \multirow{3}[2]{*}{Ours} & IL+Prediction & 45.11 & 53.38 & 23.04 & 0.479 & 0.655 & 0.283 & 5.740 & 14.81 & 37.464 &  \\
          & IL+Prediction+Planner & 7.52  & 4.51  & 73.17 & 0.552 & 1.856 & 0.328 & 1.918 & 3.772 & 9.083  \\
          & Integrated Prediction with Planning & \textbf{3.76} & 2.26  & 75.90 & 0.526 & 1.763 & 0.144 & \textbf{1.710} & \textbf{3.420} & \textbf{8.201} \\
    \midrule
    Human &       & -     & -     & -     & 0.621 & 2.067 & 0.105 & -     & -     & -     \\
    \bottomrule
    \end{tabular}%
  \label{table-2}%
\end{table*}%

\subsection{Planning Results}
Fig. \ref{figure-ours-plan} shows the qualitative planning results of our proposed framework with a motion planner. 12 representative driving scenarios are selected including passing green light, encountering red light, turning, cutting-in cars, changing lanes, cruising, and opposite oncoming traffic. In scenario (a) and (b), the ego vehicle is able to pass the green light without detention quickly. In scenario (c) and (d), when other agents are coming from the opposite direction with unclear intentions, the ego vehicle is still able to adjust and avoid collisions. In scenario (e), when the traffic light changes to red, the ego vehicle gradually slows down and comes to a stop. In scenario (f), the ego vehicle stays still during the time of the red light. Scenario (g) and (h) display the planned smooth and human-like trajectory by our method to perform a turning. Scenario (i) and (j) demonstrate the ego vehicle performing a lane-change behavior. In scenario (i), the ego vehicle change lane for faster driving, and in scenario (j), the ego vehicle change lane for turning right. Scenario (k) shows a good cruising scenario, while scenario (l) shows the case in which another agent intends to cut into the lane that the ego vehicle is currently driving. In this case, the ego vehicle chooses not to speed up and gives way to this vehicle instead. These qualitative results demonstrate that our proposed network is able to plan a smooth, safe, and compliant trajectory for the ego vehicle in complex driving scenarios.

Table \ref{table-2} shows the quantitative results of our proposed network compared with the previous state-of-the-art method \cite{huang2022differentiable}. In the aspect of imitation learning with prediction sub-task, our method outperforms the previous method. The planning errors at $3s$, $5s$, and $10s$ are reduced by $47.9\%$, $22.5\%$, and $11.3\%$ respectively, with the average progress made in each scenario being doubled. Additionally, the overall vehicle dynamics are closer to that of the expert human drivers. However, pure imitation learning with prediction sub-task still suffers poor results in the collision rate and the off-route rate. This is due to the lack of constraints involving traffic rules in pure learning-based methods and the distributional shifts in IL-based methods. When the same planner is added to refine the initial planned trajectory, the overall performance of both tasks can increase by a significant margin. Compared to the previous method, the planning errors at $5s$ and $10s$ decrease by $9.8\%$ and $10.7\%$ respectively. However, the collision rate, off-route rate, and the planning error at $3s$ of our method are slightly higher than the previous method. In the case of integrated prediction with planning, our method is trained with the planner in an end-to-end fashion (DIPP). In this case, our method still outperforms the previous method, with the planning errors at $3s$, $5s$, and $10s$ decreased by $0.9\%$,  $12.6\%$, and $12.4\%$ respectively, and a lower collision rate. Generally, our proposed network structure is able to serve as a better backbone for integrated motion prediction and planning tasks.

\section{Conclusion}
In this paper, we propose a planning-centric prediction neural network to anticipate the future states of interactive road agents as well as generate motion planning commands for the ego vehicle. A transformer-based agent-agent interaction module along the time axis is proposed to model the relationship between agents better, and the DGCNN module is utilized to extract the topological information from the map. A novel proxy way-points strategy is proposed to reduce the memory consumption of the DGCNN block. Based on these modifications, our network can provide higher prediction accuracy and better initial planning. In the future, we will use the most relevant several agents surrounding the ego vehicle as the input rather than the nearest agents.

\section*{ACKNOWLEDGMENT}
This research was supported by Yinson and Moovita Singapore.


\bibliographystyle{IEEEtran}
\bibliography{root}

\end{document}